# Forecasting Drought Using Machine Learning in California

Nan K. Li[1], Angela Chang[1], David Sherman[1]

[1]School of Information and Data Science, UC Berkeley

Corresponding author: Nan K. Li (nankli@berkeley.edu)

## Abstract

Drought is a frequent and costly natural disaster in California, with major negative impacts on agricultural production and water resource availability, particularly groundwater. This study investigated the performance of applying different machine learning approaches to predicting the U.S. Drought Monitor classification in California. Four approaches were used: a convolutional neural network (CNN), random forest, XGBoost, and long short term memory (LSTM) recurrent neural network, and compared to a baseline persistence model. We evaluated the models' performance in predicting severe drought (USDM drought category D2 or higher) using a macro F1 binary classification metric. The LSTM model emerged as the top performer, followed by XGBoost, CNN, and random forest. Further evaluation of our results at the county level suggested that the LSTM model would perform best in counties with more consistent drought patterns and where severe drought was more common, and the LSTM model would perform worse where drought scores increased rapidly. Utilizing 30 weeks of historical data, the LSTM model successfully forecasted drought scores for a 12-week period with a Mean Absolute Error (MAE) of 0.33, equivalent to less than half a drought category on a scale of 0 to 5. Additionally, the LSTM achieved a macro F1 score of 0.9, indicating high accuracy in binary classification for severe drought conditions. Evaluation of different window and future horizon sizes in weeks suggested that at least 24 weeks of data would result in the best performance, with best performance for shorter horizon sizes, particularly less than eight weeks.

## Introduction

Throughout the last few decades, the planet has experienced worsening weather conditions and increasing frequency and scale of natural disasters. Among these events, drought stands out as one of the most devastating phenomena, having significant socioeconomic and environmental consequences, particularly in regions with significant agricultural industry and limited water resources, such as California.

From 2012 to 2016, the state experienced one of its most severe and prolonged droughts in its history, which had significant residual effects, such as the loss of natural forests, native fish populations, and decreased groundwater levels (Lund et al., 2018). This drought period's dryness and heat conditions occurred with a frequency estimated to be between once in 20-1,200 years.

In 2015, the total estimated economic effect of the drought was $2.7 billion, with nearly a third of that amount stemming from crop losses, plus 21,000 total job losses (Howitt et al., 2014).

More recently, the 2020 and 2021 water years in California were the second-driest two-year periods in the history of water records and the driest two-year period since the 1976-1977 drought (PPIC, 2022). With the drought came an estimated $1.2 billion economic impact cost for 2022 alone on the agricultural industry (Medellín-Azuara et al., 2022) and an extreme series of wildfires throughout 2020. From the 2009-2018 decade, wildfires cost almost $1 billion on average annually, with up to $3.52 billion in estimated structure value loss in Butte county alone (Buechi, et al., 2021).

Drought is a slow developing natural disaster resulting from a mixture of complex factors, including water deficit and local and global weather phenomena, making drought a challenge to predict in advance (Funk and Shukla, 2020). Early detection and warning systems are key for the mitigation of negative consequences due to drought. To combat this environmentally and financially troublesome issue, several governmental and non-governmental organizations have attempted to develop tools and systems to accurately predict future drought severity. This includes the U.S. National Integrated Drought Information System, the U.S. Drought Monitor (USDM), and the Global Integrated Drought Monitoring and Prediction System, to name a few. Such methods employ the use of common drought indicators, like precipitation, temperature, streamflow, and indices, which are computed numerical representations of drought intensity (WMO et al., 2016), such as the Aridity and Crop Moisture indices.

Researchers have also attempted to utilize various time series and machine learning-based models to help improve predictive power for drought. These approaches have shown effectiveness in predicting drought intensity. Brust et al. (2021) were able to predict the USDM drought classification with MSE values of 0.0534-0.5565, or with a difference of less than one drought category, up to 12 weeks in advance using a recurrent neural network for the 2017 Northern Plains Flash Drought. Cao et al. (2023) were able to accurately predict (80-90% depending on region) the USDM drought classification using Markov chains up to 4 weeks ahead for the nation. Hameed et al. (2023) were able to develop and compare multi-month forecasting models for the Great Lakes region specifically using Extreme Learning Machine (ELM), random forest, and other hybrid models in combination with their newly developed Multivariate Standardized Lake Water Level Index (MSWI) for assessing drought.

Previous research did not seem to focus specifically on California, though some work, like Brust et al. (2021), highlighted that drought in the Western U.S. was particularly hard to predict due to the particularly slow-developing nature of drought in the region. Our goal was to focus on highlighting opportunities to predict drought intensity within the state.

Our research focused on the USDM drought classification, which depicts the intensity of droughts on a weekly basis across the country (NDMC et al., 2024) and serves as our guidelines for establishing varying drought conditions. The USDM uses a five-category system, ranging from Abnormally Dry ("D0"), a precursor to drought conditions when there is no official drought, to Exceptional Drought ("D4"), the most severe conditions. For example, for a given day, the data for a location will dictate what percent of that area is classified as D0, D1, D2, D3, D4, or none. Drought categories depict experts' assessments of conditions related to dryness and drought, including observations of how much water is available in streams, lakes, and soil compared to usual for the same time of year.

Similar to prior studies conducted by Brust and Cao, this work's goal is to determine which machine learning and/or time series-based models provided the most accurate prediction of drought intensity in California in combination with meteorological variables. Based on previous work (Nangunde et al., 2023), a long short-term memory (LSTM), convolutional neural network (CNN), and two decision tree approaches, extreme gradient boosting (XGBoost) and random forest, are presented. LSTM models have shown promising results in hydrological prediction tasks and other time series problems (Nangunde et al., 2023), as they can capture temporal relationships between features and can model non-linear relationships. The team also built a CNN model as it is well suited to capturing spatial relationships within the data, while decision tree models are able to help determine which features are most important in determining drought scores.

The models presented here were able to predict the drought intensity scores with a high degree of accuracy, with the best performing models producing F1 scores of ~90%. The models can be used by local government agencies, such as water departments, to obtain timely drought predictions. The resulting predictions can be used to implement preventative actions such as water conservation measures, agricultural planning, and disaster preparedness efforts.

## **Data**

The dataset used in this research is a public dataset available on [Kaggle](). The dataset was originally sourced from the NASA Langley Research Center (LaRC) Power Project (funded through the NASA Earth Science/Applied Science Program) and the U.S. Drought Monitor. The dataset includes daily weather measurements and weekly USDM drought score at the county-level over a 20 year period from 2000-2020. Weather variables include temperature, humidity, windspeed, precipitation, and pressure. A full list of variables can be found in the appendix (A1).

The USDM Drought Classification and the mapping to scores in the dataset is presented in Table 1.

**Table 1. Drought Score, USDM Drought Classification, and Description**

| Drought Score | USDM Drought Classification | Description |
|---|---|---|
| 0 | None | Normal or wet conditions |
| 1 | D0 | Abnormally Dry |
| 2 | D1 | Moderate Drought |
| 3 | D2 | Severe Drought |
| 4 | D3 | Extreme Drought |
| 5 | D4 | Exceptional Drought |

For this study, only counties located within California (FIPS codes in the 6000-6999 range) were utilized. Daily weather variables were averaged to produce a weekly value, resulting in 63,568 records of weekly drought scores and corresponding weekly weather variables averages for 58 CA counties spanning 1,096 weeks.

The dataset was split into training, validation, and test using a 70%, 10% and 20% split, respectively. Training data included data from 2000-2014, validation 2015-2016, and test 2017-2020. This ensured that the model encountered the full range of possible drought scores in training, though there were mostly severe drought scores (greater than 2.5) in the validation set, and no exceptional drought scores (greater than 4.5) in the test set. Detailed descriptions of modeling techniques and data engineering processes are provided in the next section.

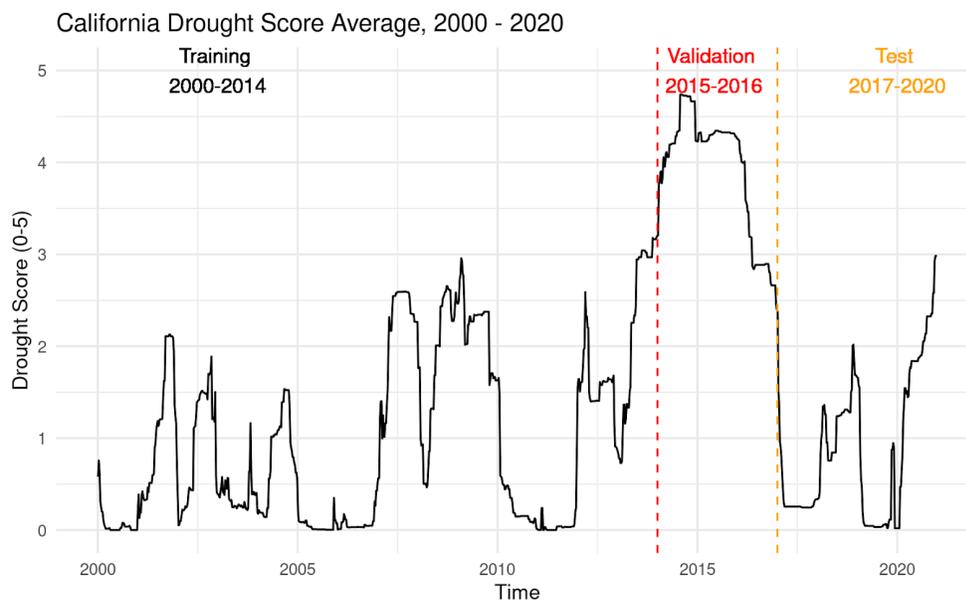

**Figure 1. Average statewide drought Score from 2000-2020**

# Models & Methods

## Model Architecture

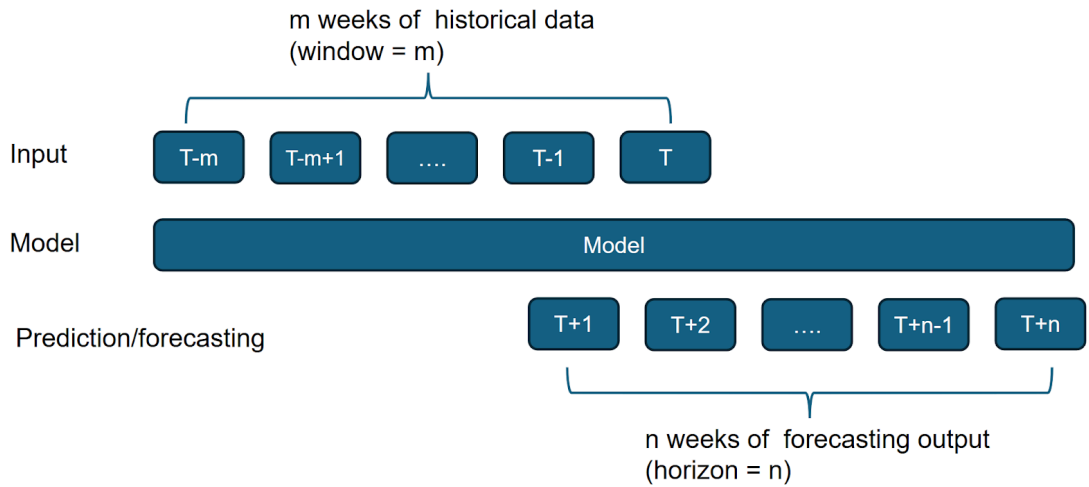

**Figure 2. Machine learning architecture for multi-step time series forecasting**

Figure 2 illustrates the machine learning architecture designed for multi-step time series forecasting, where the input incorporates historical meteorological data and drought scores from the preceding *m* weeks, while the output comprises forecasted drought scores for the subsequent *n* weeks. Here, *m* denotes the window size, and *n* represents the forecasting horizon. Initially, the window size was set to 30 weeks and the forecasting horizon to 12 weeks, based on established literature (Brust et al., 2021). Utilizing these parameters, our model leveraged historical data from the preceding 30 weeks to forecast drought scores for the subsequent 12 weeks. Notably, both the window size and forecasting horizon were tuned to evaluate the model's performance in predicting further into the future.

## Data Engineering

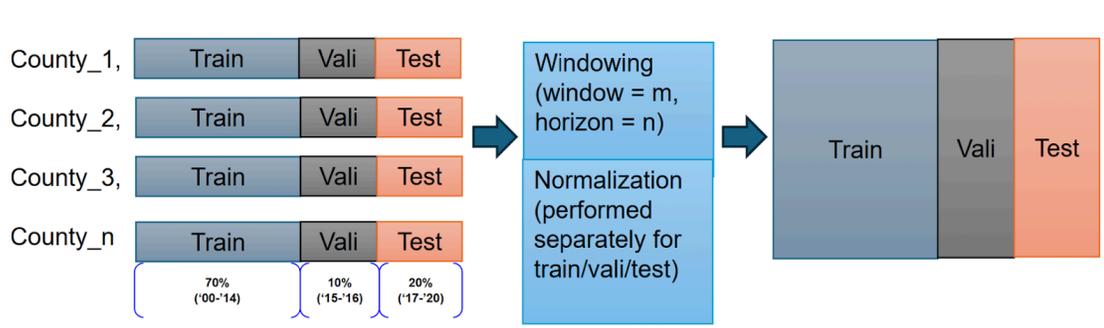

**Figure 3. Data engineering pipeline**

In the data engineering phase of our research, we assembled weekly time series data encompassing various meteorological variables and historical drought scores for each county.

Augmented with additional features such as longitude, latitude, and month to capture geospatial and temporal characteristics, the data was initially partitioned into distinct train-validation-test sets, maintaining a split ratio of 70%, 10%, and 20%, respectively. Importantly, this partitioning procedure was conducted without shuffling the data, preserving the temporal integrity of the sequences.

Subsequently, the data underwent a windowing process, wherein it was segmented into smaller, overlapping subsequences known as windows. These windows encapsulated historical data from the past $m$ weeks as features, along with corresponding drought scores for the subsequent $n$ weeks as labels. This windowing technique is instrumental in transforming the time series forecasting problem into a supervised machine learning task, enabling the model to learn from past observations to make predictions about future drought trends.

Figure 3 provides a visual representation of the data engineering pipeline employed in this study. Each train-validation-test set from individual counties was consolidated into statewide datasets, facilitating comprehensive model training and evaluation. Following consolidation, each dataset underwent normalization independently to ensure consistency in data scaling across different regions and variables.

## Baseline Model

For the baseline results for comparison, a multistep baseline or persistence model that averaged the drought score over the "Window" period was used. This baseline provided an unskilled result for comparison with our other machine learning models.

## CNN

We used a Convolutional Neural Network (CNN) architecture tailored for multistep time series forecasting tasks. The CNN model was constructed using TensorFlow's Keras API, instantiated as a sequential model. The architecture began with a 1D convolutional layer comprising 64 filters and kernel size of 3, activated using rectified linear units (ReLU). Subsequently, a max-pooling layer with a pool size of 2 was applied to downsample the feature maps. A dropout layer with a dropout rate of 0.1 was then employed for regularization to prevent overfitting. Next, a fully connected dense layer with 30 units and ReLU activation was incorporated, followed by another dropout layer with a dropout rate of 0.1. Finally, the output layer produced the predicted values for the multistep forecasting, configured with the appropriate number of units (referred to as the horizon). The model was compiled with the Adam optimizer and utilized the Mean Absolute Error (MAE) loss function. This architecture aimed to capture intricate patterns within the time series data while mitigating overfitting through dropout regularization. Hyperparameters such as the numbers of filters and the kernel size were fine-tuned to enhance model performance.

### Random Forest

The first tree-based model utilized in this study was the random forest. This model aggregates predictions from multiple decision trees, considering various combinations of samples and records. The primary parameters influencing predictive accuracy are the number of estimators, representing the total trees employed, and the maximum depth of each decision tree. This model also provides users with the ability to see which features influence the predicted value the most. In our case, historical drought score, month, Earth skin temperature, precipitation, and humidity were the top variables that influenced the predicted drought score.

After hyperparameter tuning, we found the random forest model produced optimal results when using 300 trees with a maximum depth of 4.

### XGBoost

We used one other decision tree model, Extreme Gradient Boosting (XGBoost), in this study. Like random forest, XGBoost is a decision tree ensemble model. However, when creating decision trees, XGBoost uses additive training, i.e. one tree is added at one time, as opposed to randomly creating all trees at the same time. When one tree is added at a time, the model learns from the prediction score, optimization results, and regularization term from the previous tree before creating the next tree. The model continues to create the desired number of trees, or estimators, and sums the prediction results.

After hyperparameter tuning, we found the XGBoost model produced optimal results when using 100 estimators, a max depth of 3, and a learning rate of 0.15.

### LSTM

An LSTM model tailored for multistep time series forecasting was introduced in this study. Leveraging TensorFlow's Keras API, the LSTM architecture was implemented as a sequential model. The model was structured with an initial LSTM layer comprising 150 units, followed by a dropout layer with a dropout rate set to 0.1 to mitigate overfitting. Subsequently, another LSTM layer with 75 units was incorporated, along with another dropout layer with the same dropout rate. The model concluded with a dense layer responsible for generating predictions for the multistep forecasting, configured with the appropriate number of units (referred to as the horizon). For optimization, the model was compiled with the Adam optimizer and utilized the MAE loss function to assess the disparity between predicted and actual values. Hyperparameters such as the number of units in the LSTM layers and the dropout rates in the dropout layers were fine-tuned to enhance model performance.

## **Results And Discussion**

For this study, we evaluated performance using a macro F1 score based on a binary classification of predicting "severe" drought (score of 2.5 or higher or USDM categories D2 and above) vs. non-severe drought (score of less than 2.5 or USDM categories D1 and lower). This binary classification was chosen to evaluate how well models are able to predict severe or higher drought scores in comparison to more common lower drought scores. Weighted F1 score was provided for comparison, though macro F1 was the main evaluation metric. Given that severe drought is increasing in frequency, severity, and length, the macro F1 will likely better represent a model's ability to predict future severe drought, despite low or no drought conditions at present.

The other evaluation metrics used were mean squared error (MSE) and mean absolute error (MAE) to demonstrate the difference between actual and predicted scores. MSE is calculated by averaging the squared difference between the actual and predicted values, while MAE is the average of the absolute value of the difference between the actual and predicted values.

The overall performance of the tested models is shown in Table 2.

**Table 2. Overall Model Performance**

| Model | MSE | MAE | Macro F1 | Weighted F1 |
|---|---|---|---|---|
| **Persistence (Baseline)** | 0.50 | 0.43 | **0.81** | **0.94** |
| **Random Forest** | 0.33 | 0.39 | **0.86** | **0.95** |
| **CNN** | 0.36 | 0.37 | **0.87** | **0.95** |
| **XGBoost** | 0.30 | 0.35 | **0.88** | **0.96** |
| **LSTM** | 0.32 | 0.33 | **0.90** | **0.96** |

As seen in Table 2, the best performing models according to the macro F1 score, or ability to "accurately" predict severe drought scores or higher, was the LSTM model, followed by the XGBoost model. The CNN and Random Forest model had lower but similar macro F1 scores for severe drought scores, though higher MSE and MAE values in comparison. All tested models performed better than the baseline model in all metrics.

The LSTM model had a macro F1 score of 0.90 for predicting severe drought, with a MSE of 0.32 and MAE of 0.33. The MAE and MSE indicated that the model was, on average, predicting a drought score that differed from the actual drought score of around 0.33, with drought scores ranging from 0-5. The XGBoost model performed similarly, though with a worse MAE of 0.35, or an average difference of 0.35 between the predicted and actual score.

A detailed classification report is shown in Table 3. The classification report results include precision, recall, and F1 scores for the binary classification, where 0 indicates the negative or non-severe drought category, and 1 indicates the positive or severe drought category. Macro F1 scores and the weighted F1 score is provided also.

Table 3. Models Classification Report

| Model | Class | Precision | Recall | F1 | Support | Macro F1 | Weighted F1 |
|---|---|---|---|---|---|---|---|
| **Persistence (Baseline)** | 0 | 0.95 | 0.99 | 0.97 | 110788 | | |
| | 1 | 0.86 | 0.53 | 0.66 | 13110 | | |
| | Overall | | | | | 0.81 | 0.94 |
| **Random Forest** | 0 | 0.96 | 0.92 | 0.98 | 110788 | | |
| | 1 | 0.92 | 0.63 | 0.75 | 13110 | | |
| | Overall | | | | | 0.86 | 0.95 |
| **CNN** | 0 | 0.97 | 0.98 | 0.97 | 110788 | | |
| | 1 | 0.82 | 0.71 | 0.76 | 13110 | | |
| | Overall | | | | | 0.87 | 0.95 |
| **XGBoost** | 0 | 0.96 | 0.99 | 0.98 | 110788 | | |
| | 1 | 0.90 | 0.69 | 0.78 | 13110 | | |
| | Overall | | | | | 0.88 | 0.96 |
| **LSTM** | 0 | 0.97 | 0.99 | 0.98 | 110778 | | |
| | 1 | 0.89 | 0.75 | 0.82 | 13110 | | |
| | Overall | | | | | 0.90 | 0.96 |

As demonstrated in Table 3, all models performed better at predicting non-severe or no drought in comparison to the task of predicting severe drought, with F1 for predicting non-severe drought in the 0.97-0.98 scale with high and balanced precision and recall. F1 for the positive or severe drought categories ranged from 0.66 for baseline to 0.82 for the LSTM model, with greater imbalance between precision and recall for all other models.

While precision was relatively high for all models, recall was lower for all models. High precision indicates that the model is accurately predicting true positives or actual severe drought and making few false positive predictions, i.e. inaccurately predicting severe droughts when there is none. Lower recall indicates the model is making false negative predictions, or that the model is incorrectly predicting non-severe drought scores when the actual drought scores are in the severe category. The imbalance in precision and recall is likely due to the inherent imbalance in the dataset. Severe drought is rare, making up roughly 10.5% of the test dataset. In this context, recall, or sensitivity, is likely more important. It may be more important for decision-makers that a drought prediction model can detect a severe drought in advance, as opposed to underpredicting severe drought, for more conservative decision making.

We also tested whether performance changed depending on how much past data was provided, or window size, and for the desired forecast range, or horizon. The best results from this evaluation are shown in Tables 4 and 5 for the two best performing models, LSTM and XGBoost. Full results for all horizon and window testing experiments are provided in the appendix (A2 and A3).

**Table 4. Horizon and Window Testing Results, LSTM**

| Horizon | Window(s) | Macro F1 |
| --- | --- | --- |
| 4 | 24, 30, 36, 48, 52 | 0.96 |
| 8 | 24, 30, 36, 48, 52 | 0.93 |
| 12 | 24, 30, 48 | 0.90 |
| 16 | 24 | 0.87 |

**Table 5. Horizon and Window Testing Results, XGBoost**

| Horizon | Window(s) | Macro F1 |
| --- | --- | --- |
| 4 | 30, 40, 24, 52, 12 | 0.95 |
| 8 | 40, 52, 24 | 0.912 |
| 12 | 30, 12, 40, 24, 52 | 0.88 |
| 16 | 12, 52 | 0.85 |

Results were similar for both the LSTM and XGBoost model. As expected, a shorter horizon led to better results, with macro F1 scores decreasing as the horizon increased from predicting 4 weeks out to 16 weeks after the data window. A variety of window sizes produced similar results (within 0.01 difference) in macro F1 scores, suggesting that simply increasing the window range from 12 to 24 weeks of data or from 12 to 52 weeks would not substantially improve performance. For example, a 24 week window for the LSTM model appeared to perform best at each horizon, with 30 weeks performing second best for all horizons except 16 weeks. In some cases for the XGBoost model, even using only 12 weeks of data would lead to equally good performance, while using 52 weeks did not necessarily lead to the best results.

Examining results further by using the XGBoost results as an example, the model was underpredicting drought scores. For a simpler analysis, scores were converted to 0-5 integer categories using a 0.5 threshold for each integer. For example, an actual score of 0.5 was rounded up to an integer value of one. We then compared whether the actual and predicted score categories matched, or if the prediction was correct. If the prediction was incorrect, then we then evaluated whether the predicted score was below or above the actual score. We can see in Table 6, that for an actual score of 1, when the prediction was incorrect, it predicted a smaller score 74.31% of the time. In comparison, for an actual score of 4, when the prediction was incorrect, the model predicted a smaller score 100% of the time. Or in short, the rate of underprediction for incorrect predictions increases with the actual drought score, while accuracy decreases.

**Table 6. Comparison of XGBoost Results by Score**

| Actual Score | Over Count | Over % | Under Count | Under% | Total Incorrect | Total Actual | Accuracy Rate |
|---|---|---|---|---|---|---|---|
| 0 | 10,923 | 99.95% | 6* | 0.01% | 10,929 | 62,868 | 82.62% |
| 1 | 2,638 | 25.69% | 7629 | 74.31% | 10,267 | 27,792 | 63.06% |
| 2 | 836 | 10.66% | 7010 | 89.34% | 7,846 | 20,118 | 61.00% |
| 3 | 107 | 2.73% | 3810 | 97.27% | 3,917 | 11,363 | 65.63% |
| 4 | 0 | 0% | 1347 | 100% | 1,347 | 1,747 | 22.90% |

Notes: No 5 scores in test dataset. *Negative predictions made.

In addition to evaluating model performance, we examined model feature importance in the XGBoost model. Processing feature importance from the XGBoost model suggested that the most important features were the previous drought scores, the month that the week occurred in, the earth skin temperature, precipitation, and humidity. A table of feature importance weights for the XGBoost model can be found in the appendix (A4).

The LSTM proved to be our most accurate model. To visually depict its accuracy, we plotted multiple Tableau heat maps for the actual versus predicted drought scores in Figure 4. We created a county map showcasing actual drought scores for the last 12 weeks in the dataset and score predictions for those 1-12 weeks using the previous 30 weeks of past data (the aforementioned standard prediction timeframe) for the LSTM. The weekly average actual vs. predicted drought scores can be found in Table 7. For this particular evaluation window, the model predictions were very accurate for the first 7 weeks, deviating less than 0.1 in score. Beginning in week 8, the difference between predicted and actual average scores grew substantially, particularly in weeks 10-12. From an initial scan of the week 1 vs. 12 maps, it appeared the greatest drought score discrepancies lay in counties along the Nevada border leading up to northern California; however, we found the greatest differences were due to rapid increase in drought scores over 2-3 weeks.

Week 1

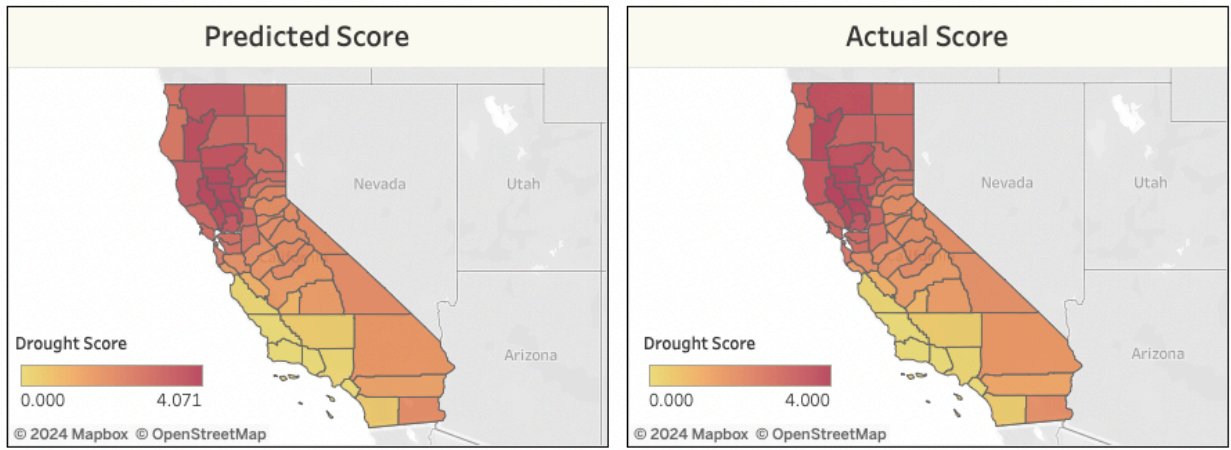

Week 12

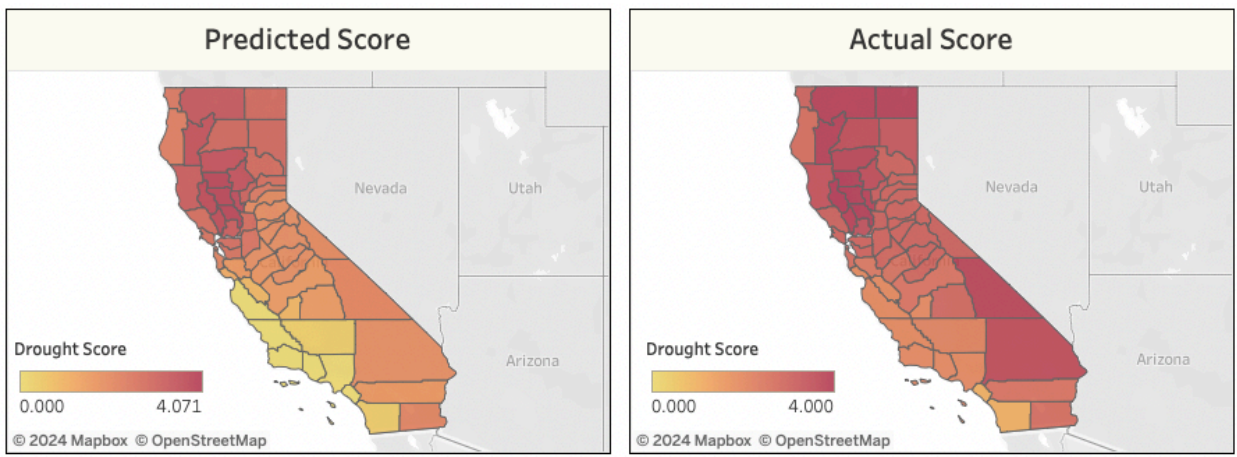

**Figure 4. Drought score maps: predicted vs actual scores for the last 12 weeks in 2020**

      Digging deeper into the weeks 10-12 time period, the maps illustrated that in San Luis Obispo, Santa Barbara, and Ventura counties, the model underpredicted by two drought categories. These counties had actual drought scores of around 2, while the model predicted no drought conditions or a score of 0. As the first 7 weeks in these counties had no drought, the model predicted a continuation of no drought conditions, despite an actual jump from a score from 0 to 2. Other counties that experienced lower model accuracies were Los Angeles and Monterey, which both had average score discrepancies of ~1.9 drought categories. Unlike in San Luis Obispo, Santa Barbara, and Ventura counties, these two all had actual and predicted drought scores greater than 0; however, the differences here could be attributed to the actual drought scores being lower than 1 for the first 8-9 weeks, before jumping to 2 in the final 3-4 weeks of our horizon. Therefore, the model only predicted drought scores of less than 0.5 for Los Angeles and Monterey due to most of the timeframe experiencing insignificant drought conditions.

      In contrast, the LSTM model worked best in areas that either regularly experience more significant drought conditions or have generally consistent drought patterns in this particular

window. For example, there were four counties with an average actual drought score of roughly 4: Lake, Yolo, Colusa, and Glenn. The LSTM model produced an average score discrepancy of 0.12 drought categories, suggesting this model would perform especially well for counties that experience more significant and consistent drought conditions.

**Table 7. Average vs. Predicted Drought Scores for Last 12 Weeks in 2020, LSTM Model**

| Week | Avg. Actual | Avg. Predicted | Avg. Discrepancy |
|---|---|---|---|
| 1 | 2.34 | 2.33 | 0.02 |
| 2 | 2.34 | 2.33 | 0.01 |
| 3 | 2.34 | 2.34 | 0.01 |
| 4 | 2.34 | 2.33 | 0.01 |
| 5 | 2.42 | 2.33 | 0.08 |
| 6 | 2.43 | 2.34 | 0.09 |
| 7 | 2.37 | 2.29 | 0.08 |
| 8 | 2.58 | 2.28 | 0.29 |
| 9 | 2.62 | 2.27 | 0.36 |
| 10 | 2.93 | 2.23 | 0.71 |
| 11 | 3.00 | 2.20 | 0.81 |
| 12 | 3.01 | 2.20 | 0.81 |

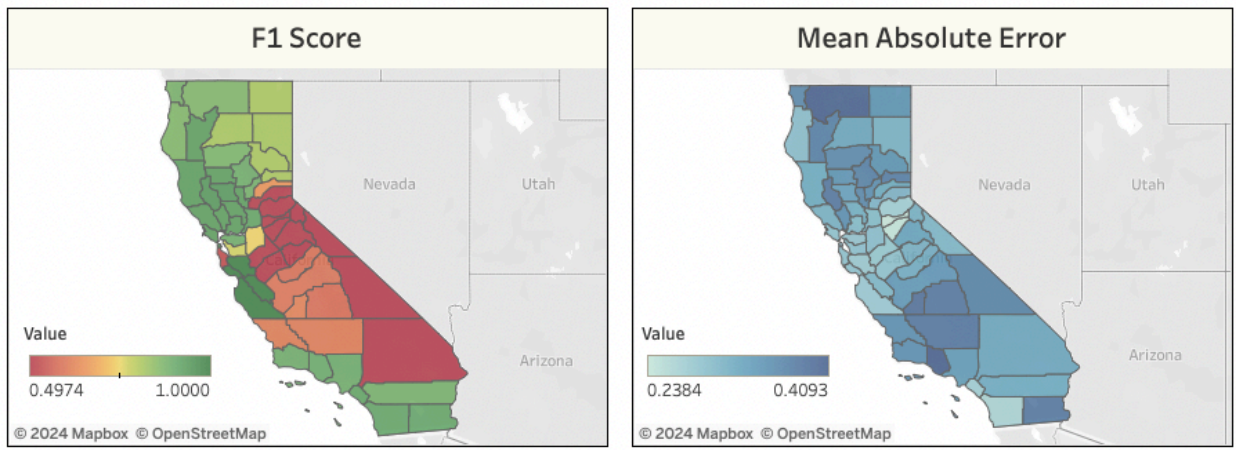

**Figure 5. Model metrics (macro F1 score and MAE) for counties**

Overall, the performance of the model exhibited variability across counties within the state, as illustrated in Figure 5. Across the full test set, F1 scores exceeded 0.86 for more than half of the counties assessed. Certain areas in the southeast of the state, notably counties along the Nevada-California border, exhibited low F1 scores. The F1 score exhibited a strong correlation with the severe drought ratio, calculated as the number of severe drought cases divided by all samples in the test data, within each county (correlation coefficient = 0.68). When the severe drought ratio fell below 0.7%, notably low F1 scores (~0.5) were observed, indicating the model's inability to predict severe drought occurrences in these areas. Additionally, F1 scores below 0.65 were commonly associated with severe drought ratios below 4% in the respective counties, suggesting a tendency for the models to underestimate severe drought occurrences. Maps illustrating MAE, seen in Figure 5, and MSE (A5 in the appendix), depict the disparity between predicted and actual drought scores, revealing consistent trends. Specifically, the models demonstrated lower prediction errors in the central region of the state.

## Conclusion

This research shows that employing an LSTM or XGBoost modeling approach yields the most effective results in predicting USDM drought classification scores in California, while outcomes using alternative common ML approaches such as CNN or random forest are marginally less accurate. Due to the imbalance of drought scores inherent in the data, that is drought being more uncommon than not, models generally underpredict extreme drought scores, particularly in areas with very low severe drought occurrences, such as by the Nevada-California border. MAE results in the range of 0.3-0.4 suggested that model predictions are deviating at generally less than half of one drought category.

Through feature importance evaluation, we found that previous drought scores and a time indication, specifically the month of the occurring score, were the most important features in guiding model predictions, in addition to Earth skin temperature, precipitation, and humidity. The importance of the previous drought score, inherent in time series problems, could also be explained by the complexity of the USDM classification score and its development based on numerous factors, including statewide conditions.

Experimenting with past window data size and future forecast horizon sizes, we found using 24 weeks or more of data resulted in reasonable performance in macro F1, 88-90%, for a 4-16 week forecast horizon. A shorter forecast horizon will result in better performance in all metrics, macro F1, MSE, and MAE, but simply increasing the amount of weeks used in training will not lead to consistent performance gains. Many of the window sizes we tested resulted in a similar performance of less than 0.01 difference in macro F1 score.

Further assessment of the best performing models, LSTM and XGBoost, indicated a common tendency to underestimate drought intensity, particularly evident for higher actual scores. As demonstrated by Tableau heat maps comparing actual and predicted drought scores over a sample 12-week period, although the LSTM showed high accuracy in the initial weeks, discrepancies between predicted and actual scores grew in later weeks. Nonetheless, the LSTM excelled in areas with persistent and significant drought conditions, suggesting its potential utility in guiding decision-making for regions susceptible to frequent drought occurrences.

Analysis of the metrics maps reveals heterogeneous model performance across counties, underscoring the intricate influence of geography, climate patterns, and water resources management in California. In this study, our predictive modeling solely relies on local weather variables and historical drought scores. However, certain regions may exhibit unique characteristics, where factors beyond local weather variables significantly impact drought conditions. In such scenarios, future work includes the integration of GIS data to enhance our comprehension and prediction of drought dynamics. GIS data offers valuable insights into various factors including soil moisture, land cover, and hydrological features, which would enrich our predictive capabilities beyond the scope of local weather variables alone. The incorporation of GIS data holds the potential to bolster the accuracy and robustness of drought forecasting systems.

When applying the same modeling approach to different areas, such as various states, challenges arise due to differences in geographical, climatic, and environmental conditions, which impact drought occurrences. Therefore, customizing or adapting model parameters and features becomes essential to achieve optimal performance across diverse regions. In California, our dataset contained over 63,000 data entries from 2000 to 2020, encompassing 58 counties. However, states with smaller geographical areas, such as Delaware, a state with only three counties, face data availability constraints. In such instances, adopting a regional modeling strategy that integrates data from neighboring areas could mitigate the scarcity of county-level data, which ensures a robust model development and assessment process.

# Appendix

## A1. Weather Variables

A full list of the weather variables in the dataset are below.

| Variable | Description |
| --- | --- |
| WS10M_MIN | Minimum Wind Speed at 10 Meters (m/s) |
| QV2M | Specific Humidity at 2 Meters (g/kg) |
| T2M_RANGE | Temperature Range at 2 Meters (C) |
| WS10M | Wind Speed at 10 Meters (m/s) |
| T2M | Temperature at 2 Meters (C) |
| WS50M_MIN | Minimum Wind Speed at 50 Meters (m/s) |
| T2M_MAX | Maximum Temperature at 2 Meters (C) |
| WS50M | Wind Speed at 50 Meters (m/s) |
| TS | Earth Skin Temperature (C) |
| WS50M_RANGE | Wind Speed Range at 50 Meters (m/s) |
| WS50M_MAX | Maximum Wind Speed at 50 Meters (m/s) |
| WS10M_MAX | Maximum Wind Speed at 10 Meters (m/s) |
| WS10M_RANGE | Wind Speed Range at 10 Meters (m/s) |
| PS | Surface Pressure (kPa) |
| T2MDEW | Dew/Frost Point at 2 Meters (C) |
| T2M_MIN | Minimum Temperature at 2 Meters (C) |
| T2MWET | Wet Bulb Temperature at 2 Meters (C) |
| PRECTOT | Precipitation (mm day-1) |

For the following two tables, A2 and A3, window size refers to the number of weeks of past data provided, while horizon refers to the desired number of weeks into the future for predictions.

## A2. Detailed window/horizon results for LSTM

| Window | Horizon | Macro F1 | MSE | MAE |
| --- | --- | --- | --- | --- |
| 12 | 4 | 0.95 | 0.14 | 0.16 |
| 12 | 8 | 0.92 | 0.24 | 0.25 |
| 12 | 12 | 0.89 | 0.34 | 0.32 |
| 12 | 16 | 0.84 | 0.44 | 0.4 |
| 24 | 4 | 0.96 | 0.12 | 0.16 |
| 24 | 8 | 0.93 | 0.22 | 0.25 |
| 24 | 12 | 0.9 | 0.32 | 0.34 |
| 24 | 16 | 0.87 | 0.41 | 0.39 |
| 30 | 4 | 0.96 | 0.12 | 0.16 |
| 30 | 8 | 0.93 | 0.23 | 0.25 |
| 30 | 12 | 0.9 | 0.33 | 0.33 |
| 30 | 16 | 0.83 | 0.45 | 0.42 |
| 36 | 4 | 0.96 | 0.13 | 0.17 |
| 36 | 8 | 0.93 | 0.22 | 0.25 |
| 36 | 12 | 0.89 | 0.36 | 0.36 |
| 36 | 16 | 0.85 | 0.42 | 0.41 |
| 48 | 4 | 0.96 | 0.14 | 0.18 |
| 48 | 8 | 0.93 | 0.24 | 0.28 |
| 48 | 12 | 0.9 | 0.35 | 0.35 |
| 48 | 16 | 0.86 | 0.48 | 0.44 |
| 52 | 4 | 0.96 | 0.14 | 0.18 |

| Window | Horizon | Macro F1 | MSE | MAE |
|---|---|---|---|---|
| 52 | 8 | 0.93 | 0.25 | 0.29 |
| 52 | 12 | 0.89 | 0.39 | 0.39 |
| 52 | 16 | 0.86 | 0.48 | 0.43 |

## A3. Detailed window/horizon results-XGBoost

| Window | Horizon | Macro F1 | MSE | MAE |
|---|---|---|---|---|
| 12 | 4 | 0.95 | 0.12 | 0.19 |
| 12 | 8 | 0.91 | 0.21 | 0.27 |
| 12 | 12 | 0.88 | 0.28 | 0.33 |
| 12 | 16 | 0.85 | 0.36 | 0.39 |
| 24 | 4 | 0.95 | 0.11 | 0.18 |
| 24 | 8 | 0.92 | 0.20 | 0.27 |
| 24 | 12 | 0.88 | 0.28 | 0.34 |
| 24 | 16 | 0.84 | 0.37 | 0.40 |
| 30 | 4 | 0.95 | 0.12 | 0.19 |
| 30 | 8 | 0.91 | 0.21 | 0.28 |
| 30 | 12 | 0.88 | 0.30 | 0.35 |
| 30 | 16 | 0.84 | 0.38 | 0.41 |
| 40 | 4 | 0.95 | 0.12 | 0.20 |
| 40 | 8 | 0.92 | 0.22 | 0.30 |
| 40 | 12 | 0.88 | 0.32 | 0.37 |
| 40 | 16 | 0.84 | 0.41 | 0.44 |
| 52 | 4 | 0.95 | 0.13 | 0.22 |
| 52 | 8 | 0.92 | 0.23 | 0.31 |
| 52 | 12 | 0.88 | 0.33 | 0.39 |
| 52 | 16 | 0.85 | 0.43 | 0.45 |

## A4. XGBoost Feature Importance Weights

Temperature, water/humidity, and wind speed related variables are indicated by color.

| Variable | Importance Weight |
|---|---|
| Drought Score (0-5) | 0.272987 |
| Month of the Score Recording (1-12) | 0.212558 |
| Earth Skin Temperature (C) | 0.061775 |
| Precipitation (mm day-1) | 0.051974 |
| Specific Humidity at 2 Meters (g/kg) | 0.046246 |
| Temperature Range at 2 Meters (C) | 0.040055 |
| Minimum Temperature at 2 Meters (C) | 0.037579 |
| Maximum Temperature at 2 Meters (C) | 0.036907 |
| Temperature at 2 Meters (C) | 0.034056 |
| Dew/Frost Point at 2 Meters (C) | 0.032309 |
| Wet Bulb Temperature at 2 Meters (C) | 0.031222 |
| Surface Pressure (kPa) | 0.021763 |
| Wind Speed Range at 10 Meters (m/s) | 0.02155 |
| Wind Speed at 10 Meters (m/s) | 0.020136 |
| Maximum Wind Speed at 10 Meters (m/s) | 0.018703 |
| Maximum Wind Speed at 50 Meters (m/s) | 0.018349 |
| Wind Speed at 50 Meters (m/s) | 0.014036 |
| Wind Speed Range at 50 Meters (m/s) | 0.009933 |
| Minimum Wind Speed at 10 Meters (m/s) | 0.008525 |
| Minimum Wind Speed at 50 Meters (m/s) | 0.007471 |
| Longitude | 0.001097 |
| Latitude | 0.000768 |

## A5. County heatmap for MSE

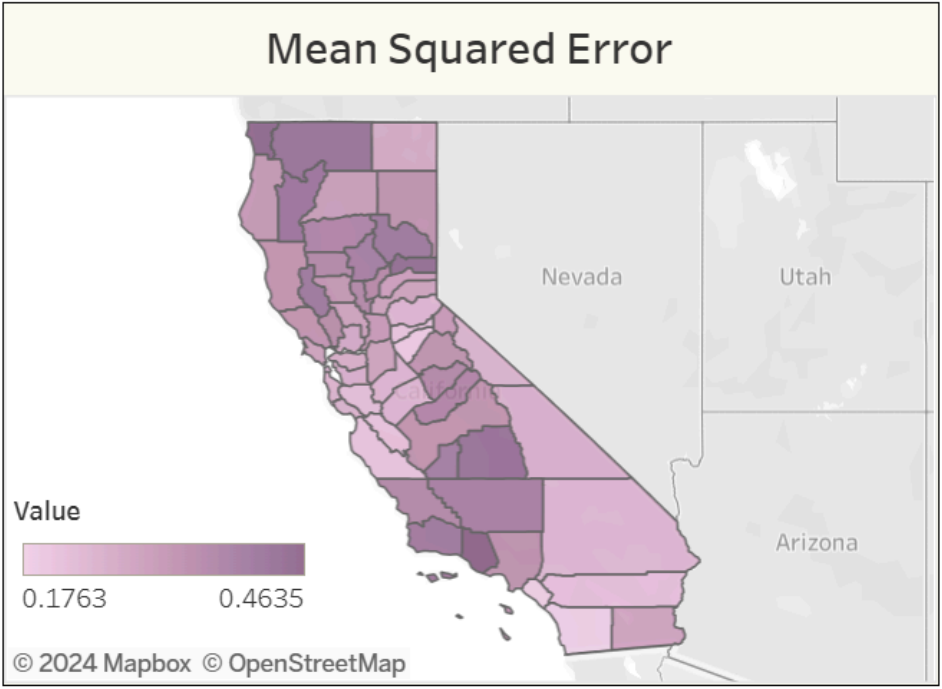